\documentclass[sigconf]{acmart}

\usepackage{amsmath,amsfonts,bm}

\def\eqref#1{equation~\ref{#1}}

\def\1{\bm{1}}

\DeclareMathAlphabet{\mathsfit}{\encodingdefault}{\sfdefault}{m}{sl}
\SetMathAlphabet{\mathsfit}{bold}{\encodingdefault}{\sfdefault}{bx}{n}

\newcommand{\E}{\mathbb{E}}

\DeclareMathOperator*{\argmin}{arg\,min}

\AtBeginDocument{%
  \providecommand\BibTeX{{%
    \normalfont B\kern-0.5em{\scshape i\kern-0.25em b}\kern-0.8em\TeX}}}

\acmConference[AdvML '20]{AdvML '20: Workshop on Adversarial Learning Methods for Machine Learning and Data Mining}{August 24, 2020}{San Diego, CA}
\acmBooktitle{AdvML '20: Workshop on Adversarial Learning Methods for Machine Learning and Data Mining, August 24, 2020, San Diego, CA}

\begin{document}

\title{Improving LIME Robustness with Smarter \\ Locality Sampling}

\author{Sean Saito}
\affiliation{%
  \institution{SAP}
  \country{Singapore}
}
\email{sean.saito@sap.com}

\author{Eugene Chua}
\affiliation{%
  \institution{UCSD}
  \city{La Jolla, CA}
  \country{USA}}
  \email{eychua@ucsd.edu}

\author{Nicholas Capel}
\affiliation{%
  \institution{Gallenco Science}
  \country{Singapore}}
  \email{nicholas.rj.capel@gmail.com}

\author{Rocco Hu}
\affiliation{%
 \institution{SAP}
 \country{Singapore}}
 \email{rocco.hu@sap.com}

\renewcommand{\shortauthors}{S. Saito et al.}

\begin{abstract}
Explainability algorithms such as LIME have enabled machine learning systems to adopt transparency and fairness, which are important qualities in commercial use cases. However, recent work has shown that LIME's naive sampling strategy can be exploited by an adversary to conceal biased, harmful behavior. We propose to make LIME more robust by training a generative adversarial network to sample more realistic synthetic data which the explainer uses to generate explanations. Our experiments demonstrate an increase in accuracy across three real-world datasets in detecting biased, adversarial behavior compared to vanilla LIME. This is achieved while maintaining comparable explanation quality, with up to 99.94\% in top-1 accuracy in some cases.
\footnote{Code for our experiments can be found at https://github.com/seansaito/Faster-LIME}


\end{abstract}

\ccsdesc[500]{Computing methodologies~Supervised learning by classification}
\ccsdesc[100]{Human-centered computing~Systems and tools for interaction design}
\keywords{explainability, robustness, adversarial machine learning}

\maketitle

\section{Introduction}
\label{sec:introduction}


Explainability is a topic of growing interest especially in applications where trust and transparency are requirements. Several methods exist for explaining the decisions of an otherwise black-box machine learning model, including the Locally Interpretable Model-Agnostic Explanation (LIME) algorithm \cite{Ribeiro2016quotWhyClassifier} and SHAP \cite{LundbergAPredictions}.

Recent work in adversarial machine learning has shown that even these explainability algorithms are vulnerable to adversarial attacks. The objectives of the attacks vary, including extracting a high-fidelity copy of the black-box model (\cite{Jagielski2019HighNetworks}), identifying training data membership of individual data points (\cite{Shokri2019PrivacyModels}, \cite{NasrComprehensiveLearning}), or concealing harmful, biased behavior of black-box models on categories like race, sex, and religion (\cite{Slack2019FoolingMethods}).

In this work, we propose a method which addresses the last type of attack. In particular, we propose using the Conditional Tabular GAN (CTGAN) model \cite{Xu2019ModelingGAN} to generate more realistic synthetic data for querying the model to be explained. Our experiments involving both black-box and white-box adversarial attacks demonstrate that our model can more accurately detect biased behavior as compared to the vanilla LIME model. To the best of our knowledge, this is one of the first attempts at making LIME more robust against adversarial attacks.



\section{Overview of LIME}
\label{sec:about_lime}

Given some target model $f$ to be explained, LIME produces an explanation $e$ for it by optimizing for both \textit{local model fidelity} and \textit{complexity}. For some prediction $f(x)$, LIME produces some explanation $e(x)$ which optimizes the following:

\vspace{-2.5mm}
\begin{equation}
    \argmin_e \mathcal{L}(f, e, \pi_x) + \Omega(e)
\end{equation}



Local model fidelity $\mathcal{L}$ here refers to how faithful the explanation is to the model being explained, \textit{f}, in a locality measured by $\pi_{x}$ around a specific prediction $f(x)$. In practice, $e$ is typically a generalized linear model and $\mathcal{L}$ is usually defined as a locally-weighted squared loss: 

\vspace{-2mm}
\begin{equation}
    \mathcal{L} = \sum_{z, z' \in \mathcal{Z}} \pi_x(z) (f(z) - e(z'))^2
\end{equation}
\vspace{-2mm}

\noindent where $z, z' \in \mathcal{Z}$ refer to synthetic data sampled around $x$ and its binary representation, and $\pi_x(z) = exp(\frac{-D(x, z)^2}{\sigma^2})$. $D$ represents some appropriate distance measure according to the data domain (cosine distance for text data and $L_2$ distance for images). LIME only relies on query-access to the black-box model for constructing pairs of $(z, f(z))$, thereby establishing \textit{model agnosticism}.









\begin{figure}[h]
  \centering
  \includegraphics[width=0.85\linewidth]{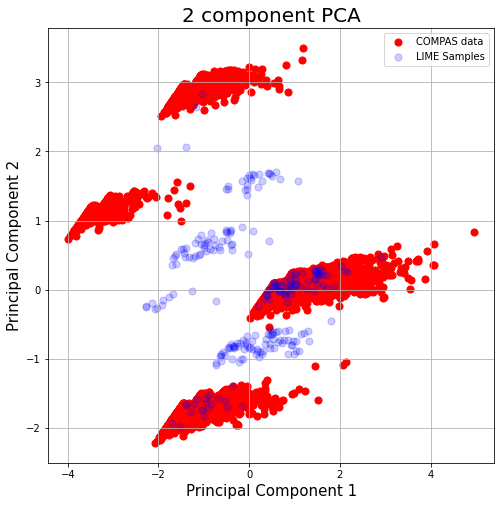}
  \caption{2-D PCA projection of real data and synthetic data generated by LIME}
  \Description{Figure showing poor overlapping of dataset and vanilla sampled artificial data. The points are clearly not overlapping.}
  \label{fig:pcaperturbed}
\end{figure}

\subsection{Attack on LIME}
\label{subsec:attack_on_lime}

A notable drawback with LIME's sampling method -- and the core problem we seek to address in this work -- is that the synthetic samples are typically generated by sampling from a unit Gaussian, $X \sim \mathcal{N}(0,\,1)\,$ around the instance to be explained. Recently, Slack et al. \cite{Slack2019FoolingMethods} have identified that this leads to the creation of explanations based on identifiably out-of-distribution (\textit{OOD}) synthetic data, as shown in Figure \ref{fig:pcaperturbed}. They subsequently proposed an adversarial attack -- which we subsequently refer to as the Fooling-LIME attack -- that exploits this vulnerability and hides biased behavior from explainers like LIME.

In particular, suppose that an adversary has a biased classifier $a$ which he/she wants to hide from the auditor who uses explainability algorithms to detect biased, harmful behavior. The adversary first constructs a critic $c$ which predicts how realistic a given data point is. This can be done by creating a dataset comprised of real data and synthetic samples generated by LIME, and then training $c$ on this binary classification task. A scaffolding function $s$ is then used to route data points based on predictions of $c$:

\begin{equation}
    s(x)= 
\begin{cases}
    a(x),& \text{if } c(x) \geq 0.5\\
    \psi(x),              & \text{otherwise}
\end{cases}
\end{equation}

In other words, if $c$ determines $x$ to be an \textit{OOD} sample, i.e. coming from an explainer query, $s$ routes $x$ to $\psi$, a model which exhibits unbiased behavior. The explainer, given prediction behavior coming from $\psi$ rather than $a$, would be fooled into determining that the adversary is unbiased and innocuous.

\section{Improving the robustness of LIME}
\label{sec:method}
In this section, we introduce our proposed methods for improving the robustness of LIME.

\subsection{Smarter locality sampling}
\label{subsec:sampling}

We propose to employ a sampling model which learns to generate realistic data. More specifically, we would like a data sampler $g$ that optimizes the following objective:

\begin{equation}
\begin{aligned}
    & {\text{maximize}}
    & & c(g(x)) \\
    & \text{s.t.}
    & & \left | x - g(x) \right | \leq \delta \\
\end{aligned}
\end{equation}

In other words, the goal of $g$ is to "fool" the adversarial critic $c$ into thinking that the samples generated by $g$ are indeed real. $x = \{c_1, \dots c_n, d_1, \dots, d_m\}$ is a row from some tabular data, which represents a concatenation of continuous variables $C_1, \dots, C_n$ and discrete variables $D_1, \dots, D_m$. This optimization problem is similar to that of generative adversarial networks, and we thus use the Conditional Tabular GAN (CTGAN) model \cite{Xu2019ModelingGAN} as $g$. CTGAN overcomes the limitations of the vanilla GAN by facilitating sampling from non-Gaussian and multi-modal distributions, creating more realistic tabular data. It generates data by taking in two inputs - a noise vector $z$ sampled from a standard normal distribution and a conditional vector $m$ that selects categories from the discrete variables which influences the output of the generator. As mentioned in \cite{Xu2019ModelingGAN}, we use WGAN loss to train CTGAN:

\begin{equation}
  L = \E \left[d(g(z, m))\right] - \E \left[d(x) \right] - \lambda \E \left[ \left( \left \|  \nabla_{\hat{x}} d(\hat{x}) \right \|_2 - 1 \right)^2 \right]  
\end{equation}

where the last term is the gradient penalty from \cite{Gulrajani2017ImprovedGANs} which enforces Lipschitz smoothness of the model weights. During inference, instead of generating synthetic data from a randomly sampled conditional vector, we condition CTGAN with the categorical values of $x$, the given data instance to be explained:

\begin{equation}
    \Tilde{X} = g(z, m_x)
\end{equation}

where $m_x$ is a concatenation of binary mask vectors $m_{d_1}, \dots m_{d_m}$, each indicating the categorical values of $x$ in one-hot form:

\begin{equation}
\begin{aligned}
    & m_{d_i}^{k} = &&
\begin{cases}
    1,& \text{if } k = x_{d_i}\\
    0,              & \text{otherwise}
\end{cases} \text{for } k = 1, \dots, |D_i| \\
    & m_x  && = m_{d_1} \oplus m_{d_2} \cdots \oplus m_{d_n} \\
\end{aligned}
\end{equation}

Using the actual categorical values of $x$ conditions $g$ to generate synthetic data that is more local with respect to $x$. We further enhance the synthetic neighborhood by utilizing the discriminator $d(x)$ obtained through the CTGAN training process to prune unrealistic samples:

\begin{equation}
\Tilde{X}_{filtered} = \{x | x \in \Tilde{X}, d(x) \geq \tau\}
\end{equation}

where $\tau$ is a threshold for filtering out unrealistic samples. The final synthetic samples $X_{filtered}$ are then used to optimize a generalized linear model, with the coefficients representing the attributions of each feature to the given model prediction. Table \ref{table:compas_saples} displays sample COMPAS data generated by CTGAN. We subsequently refer to our proposed method as CTGAN-LIME.

\begin{table}[]
    \label{table:sample_ctgan}
    \centering
    \caption{Sample COMPAS data generated by CTGAN}
    \scalebox{0.9}{
    \begin{tabular}{rrrrr}
    \toprule
     age &  two\_year\_recid &  priors\_count &  length\_of\_stay &  race  \\
    \midrule
      35 &               0 &             12 &             541 &     1     \\
      20 &               1 &             2 &             248 &     0           \\
      25 &               1 &            2 &              31 &     1        \\
      25 &               0 &             4 &             365 &     0          \\
      43 &               0 &             9 &              62 &     1         \\
    \bottomrule
    \end{tabular}
    }
    \label{table:compas_saples}
\end{table}

\section{Threat model}
\label{sec:threat_model}

We now provide context to our work by elaborating the threat model under which we conduct our experiments.

\subsection{Adversarial objectives}
\label{subsec:objectives}

We suppose an adversary whose goal is to deploy an adversarial model $a$ into some productive landscape. $a$ is adversarial in that it produces biased, unfair predictions based on sensitive features of the data. For example, in the COMPAS dataset, where the task is to predict the likelihood of recidivism for a given convict, $a$ could be implemented as:

\begin{equation}
    a(x)= 
\begin{cases}
    1,& \text{if } x_{race} = \textit{"African American"}\\
    0,              & \text{otherwise}
\end{cases}
\end{equation}

In the process of deploying the model, the adversary may be subject to an auditory process. We assume that the implementation of $a$ is hidden from the auditor to preserve intellectual property. Hence the auditor may use some explainer $e$ to query the model and discover any biased, harmful behavior. The objective of the adversary is then to fool $e$ into thinking that $a$ is innocuous.

\subsection{Adversarial settings}
\label{subsec:adv_setting}

We experiment with two adversarial settings, namely \textit{black-box} and \textit{white-box}. In the former, the adversary does not have access to the explainer $e$ nor its implementation; it only receives queries from $e$. In our experiments, we attack CTGAN-LIME using the original Fooling-LIME attack, which trains its critic $c$ based on synthetic samples generated by the LIME strategy, to measure robustness of CTGAN-LIME against the prevailing adversarial attack.

However, to further fortify our investigation, we also conduct experiments under the white-box setting, where we assume that the adversary now has increased access to $e$, including its specifications, implementations, and even its parameters. In our white-box experiments, the attacker has full access to the CTGAN generator of CTGAN-LIME. In other words, the attacker generates training data for $c$ \textit{from the same sampler} which the explainability model uses to generate explanations. This is a strictly stronger adversarial setting than the black-box counterpart and provides a stronger evaluation of robustness.

\subsection{Evaluation}
\label{subsec:eval}

We measure the robustness of an explainability algorithm based on \textit{top-k accuracy}, or the proportion of explanations which correctly identify the sensitive feature (e.g. race, gender) to be among the top $k$ contributing features for the adversarial model $a$. In other words, this represents how well the explainability model is able to detect biased behavior.




\begin{table*}[!ht]
\caption{Top-$k$ accuracy of explainers against the black-box Fooling-LIME attack with varying values of $k$}
\label{table:black_box_accuracy}
\begin{tabular}{lccccccccc}
\toprule
            & \multicolumn{3}{c}{COMPAS} & \multicolumn{3}{c}{German Credit} & \multicolumn{3}{c}{Communities} \\
Explainer / k & 1       & 3       & 5      & 1         & 3         & 5         & 1        & 10        & 30       \\ \hline
LIME &    0.00     &     42.71    &    70.90    &    0.00 &    47.20 & 65.60 & 0.00 & 9.22&  35.47        \\
CTGAN-LIME & 99.74 & 99.55 & 99.68 & \textbf{61.20} & 59.20 & 61.60 & 2.20 & 31.26 & 48.50        \\
CTGAN-LIME with $d(x)$ & \textbf{99.94} & \textbf{100.00}  & \textbf{100.00} & 60.80 & \textbf{65.60} & \textbf{67.60} & \textbf{7.01} & \textbf{83.97} &   \textbf{95.99}    \\ \bottomrule
\end{tabular}
\end{table*}

\begin{table*}[!ht]
\caption{Top-$k$ accuracy of explainers against white-box Fooling-LIME attack with varying values of $k$}
\label{table:white_box_accuracy}
\begin{tabular}{lccccccccc}
\toprule
            & \multicolumn{3}{c}{COMPAS} & \multicolumn{3}{c}{German Credit} & \multicolumn{3}{c}{Communities} \\
Explainer / k & 1       & 3       & 5      & 1         & 3         & 5         & 1        & 10        & 30       \\ \hline
LIME        &    0.00     &    38.43     &    69.09 & 0.00 & 44.80 &     64.80 & 0.00 & 8.82 & 36.07  \\
CTGAN-LIME  & 82.70 &  88.14 & 88.74 & 16.80 & 52.40 & 56.80 & 0.60 & 10.82 & 25.45        \\
CTGAN-LIME with $d(x)$ &  \textbf{84.33}  &  \textbf{96.57}  & \textbf{96.81}  & \textbf{28.80} & \textbf{58.80} &  \textbf{65.00} & \textbf{6.21} & \textbf{83.17}&   \textbf{94.59}   \\ \bottomrule
\end{tabular}
\end{table*}

\section{Experiment details}
\label{subsec:experiment_details}

In this section we elaborate our experimental details and settings. We measure robustness across the following three datasets:

\begin{itemize}
    \item \textbf{COMPAS recidivism} \cite{larson2016we}: A dataset where the task is to predict whether a given criminal defendant will commit another offense. The sensitive feature used by the biased classifier $a$ is \textit{Race}.
    \item \textbf{German credit} \cite{asuncion2007uci}: A dataset which classifies how good or bad a given loan applicant is. The sensitive feature is \textit{Gender}, and the biased classifier predicts "good" if \textit{Gender = "male"} and "bad" if otherwise.
    \item \textbf{Communities and crime} \cite{redmond2002data}: A dataset which measures various socio-economic and law-enforcement variables of communities in the US. As done in \cite{Slack2019FoolingMethods}, we convert the dataset into a classification task whereby a community is labelled "high-crime" if the violent crime rate is above the median and "low-crime" otherwise. We designate the \textit{white population proportion} as the sensitive attribute. The biased classifier predicts "high-crime" if $\textit{white population proportion} < 0.5$ and "low-crime" if otherwise.
\end{itemize}

Explanations are generated using the test set of each dataset. For all experiments, we use a Random Forest classifier as the black-box model with which we generate explanations. For each dataset, we vary the number of top features $k$ to be returned by the explainer. Both COMPAS and German credit data contain 9 features each, hence we choose $k=\{1,3,5\}$. The communities and crime dataset contains 100 features, and hence the choice $k=\{1, 10, 30\}$ is more appropriate. 



\section{Results}
\label{subsec:results}

In this section, we present the results of each of our experiments. Section \ref{subsec:robustness_results} compares the robustness of LIME and CTGAN-LIME. Section \ref{subsec:explainer_results} compares the quality of explanations of each explainer, and Section \ref{subsec:qualitative} presents a qualitative comparison of the synthetic samples generated by the respective methods.

\subsection{Robustness comparison}
\label{subsec:robustness_results}





Tables \ref{table:black_box_accuracy} and \ref{table:white_box_accuracy} refer to the accuracies of the explainability model on predictions generated by the black-box and white-box Fooling-LIME attacks respectively. For each dataset, we measure accuracy for different values of $k$, or the number of top features generated by the explainer. 

CTGAN-LIME achieves higher accuracy than LIME for each dataset, indicating that it can better detect biased behavior of the adversarial model across different datasets. In the white-box attack setting, the attacker has access to the CTGAN model and its parameters, thereby enabling the attacker to execute a more effective attack. Indeed, Table \ref{table:white_box_accuracy} displays a decrease in accuracy of the CTGAN-LIME model. However, we also note that the accuracy CTGAN-LIME is still higher than that of LIME, indicating increased robustness overall; in fact, the Fooling-LIME attack using the CTGAN sampler is slightly more effective than the original attack. Finally, we observe that the discriminator $d(x)$ which filters low-quality samples helps improve the accuracy of CTGAN-LIME further.

\subsection{Explainer comparison}
\label{subsec:explainer_results}

Beyond robustness, we also measure the quality of explanations of CTGAN-LIME and compare them to those of LIME. To this end, we rely on the precision metric proposed in \cite{RibeiroAnchors:Explanations}:

\begin{equation}
precision(E) = \E_{Q(x'|E)} \left [ \1_{f(x)=f(x')} \right ]
\end{equation}

where $E$ is the explanation and $Q(x'|E)$ is a set of samples $x' \in X$ which satisfy the predicates in $E$. What precision aims to measure is the agreement between the model's prediction $f(x)$ and the model's prediction on other samples which are similar to $x$ according to the features identified as influential in $E$. Table \ref{table:precision} reports the precision of CTGAN-LIME on each dataset and shows that CTGAN-LIME can achieve comparable precision to LIME. This suggests that the quality of explanations by CTGAN-LIME is on par with that of LIME.

\begin{table}[ht]
\centering
\caption{Precision on each dataset}
\label{table:precision}
\scalebox{1.0}{
\begin{tabular}{lcccc}
\toprule
            & COMPAS & German Credit & Communities \\
\midrule
LIME  &  71.49  & 74.58  & 82.87 \\
CTGAN-LIME & 73.27 & 77.89 & 80.64 \\
\bottomrule
\end{tabular}
}
\label{table:acc}
\end{table}

\subsection{Qualitative analysis}
\label{subsec:qualitative}

\begin{figure}[h]
  \centering
  \includegraphics[width=0.85\linewidth]{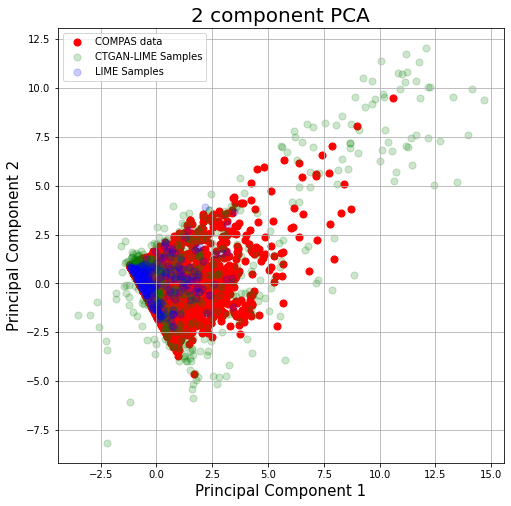}
  \caption{PCA of numerical features for COMPAS dataset}
  \Description{Figure showing better overlap of robust sampled data and data as compared to the vanilla sampled data}
  \label{fig:pcacompas}
\end{figure}

In Figure \ref{fig:pcacompas}, the numerical data of the COMPAS dataset is projected onto the first and second principal component. The robust samples drawn from CTGAN (green) more realistically represent the true data (red) than the data generated by the vanilla sampler (blue). 
The vanilla samples are clustered in a small area, and the scale is roughly equal in the direction of both principal axes. This agrees with intuition, as the data is sampled from a unit Gaussian ball. However, the true underlying data manifold is clearly long tailed and not Gaussian. The CTGAN sampler is hence better able to represent the underlying data distribution and improve robustness.

\section{Conclusion and future work}

We have shown empirical results suggesting that utilizing a generative adversarial network like CTGAN to sample synthetic data for generating explanations can improve robustness towards adversarial attacks. We hope this work will inspire additional efforts towards making explainability algorithms like LIME more robust and reliable. In particular, future work would focus on further conducting empirical evaluations as well as establishing theoretical justifications and bounds for the robustness of explanations.

\section*{Acknowledgements}

We would like to acknowledge and thank the Institute of Practical Ethics at University of California San Diego for supporting our project.

\bibliographystyle{ACM-Reference-Format}
\bibliography{references, other}

\end{document}